\documentclass{article}
\usepackage{ijcai22}

\usepackage{times}
\usepackage{soul}
\usepackage{url}
\usepackage[hidelinks]{hyperref}
\usepackage[utf8]{inputenc}
\usepackage[small]{caption}
\usepackage{graphicx}
\usepackage{amsmath}
\usepackage{amsthm}
\usepackage{booktabs}
\usepackage{algorithm}
\usepackage{algorithmic}
\usepackage{url}            
\usepackage{amsfonts}       
\usepackage{nicefrac}       
\usepackage{microtype}      
\usepackage{xcolor}         
\usepackage{subfigure}
\usepackage{stmaryrd}
\usepackage{enumitem}
\usepackage{comment}
\usepackage{amsthm}
\usepackage{bm}
\usepackage{makecell}
\usepackage{pgf}

\usepackage{ragged2e}

\newtheorem{theorem}{Theorem}

\title{Composing Neural Learning and Symbolic Reasoning with an Application to Visual Discrimination}

\author{
Adithya Murali$^1$\footnote{Contact Authors}\and
Atharva Sehgal$^{2\,*}$\and
Paul Krogmeier$^1$\And
P. Madhusudan$^1$\\
\affiliations
$^1$Department of Computer Science, University of Illinois at Urbana-Champaign\\
$^2$Department of Computer Science, University of Texas at Austin\\
\emails
adithya5@illinois.edu, atharvas@utexas.edu, paulmk2@illinois.edu, madhu@illinois.edu
}

\begin{document}
\newcommand{\paul}[1]{\textcolor{red}{#1}}
\newcommand{\adithya}[1]{}
\newcommand{\madhu}[1]{\textcolor{green}{#1}}
\newcommand{\atharva}[1]{\textcolor{orange}{#1}}

\newcommand{\figref}[1]{Figure~\ref{fig:#1}}
\newcommand{\figlabel}[1]{\label{fig:#1}}
\newcommand{\secref}[1]{Section~\ref{sec:#1}}
\newcommand{\seclabel}[1]{\label{sec:#1}}
\newcommand{\defref}[1]{Definition~\ref{def:#1}}
\newcommand{\deflabel}[1]{\label{def:#1}}

\newtheorem*{theorem*}{Theorem}
\newtheorem*{lemma*}{Lemma}
\newtheorem{definition}{Definition}
\newtheorem*{definition*}{Definition}
\newtheorem*{proof*}{Proof}

\newcommand{\anonsite}{\hyperlink{https://anonymousocean.github.io/}{\textcolor{blue}{https://anonymousocean.github.io/}}}
\newcommand{\clevr}{{\sc CLEVR}}
\newcommand{\bfclevr}{{\sc \textbf{CLEVR}}}
\newcommand{\oddone}{{\sc OddOne}}
\newcommand{\bfoddone}{{\sc \textbf{OddOne}}}

\newcommand{\hidecite}[1]{}

\newcommand{\ijcaicut}[1]{}

\newcommand{\clevrbase}{15}
\newcommand{\clevrvariants}{25}
\newcommand{\natbase}{20}
\newcommand{\clevrnum}{825}
\newcommand{\altnum}{1872}
\newcommand{\natnum}{3864}
\newcommand{\gqanum}{5000}
\newcommand{\totalnum}{11600}
\maketitle

\begin{abstract}
We consider the problem of combining machine learning models to perform higher-level cognitive tasks with clear specifications. We propose the novel problem of Visual Discrimination Puzzles (VDP) that requires finding interpretable discriminators that classify images according to a logical specification. Humans can solve these puzzles with ease and they give robust, verifiable, and interpretable discriminators as answers. We propose a compositional neurosymbolic framework that combines a neural network to detect objects and relationships with a symbolic learner that finds interpretable discriminators. We create large classes of VDP datasets involving natural and artificial images and show that our neurosymbolic framework performs favorably compared to several purely neural approaches. 
\end{abstract}

\section{Introduction}

Deep learning has made significant strides in solving specialized
tasks, especially in areas such as vision and
NLP~\cite{Goodfellow16}.  In this paper we 
study how these specialized models 
can be \emph{composed and integrated} into solutions for high-level
tasks with clear specifications. We are especially interested in
solutions for settings without a lot of data.

We believe the problem of integrating learned models is important and
can have many applications. For example, a robot may need to formulate
a complex plan that utilizes pretrained vision components to detect
objects and prediction of human behavior. 
It is challenging to utilize pretrained components to achieve a high-level task.
In this paper, we investigate neurosymbolic techniques for higher level symbolic reasoning using neural components for such problems. 

%

%


\subsection{Visual Discrimination Puzzles}
\begin{figure}
\begin{center}
    \includegraphics[width=0.48\textwidth]{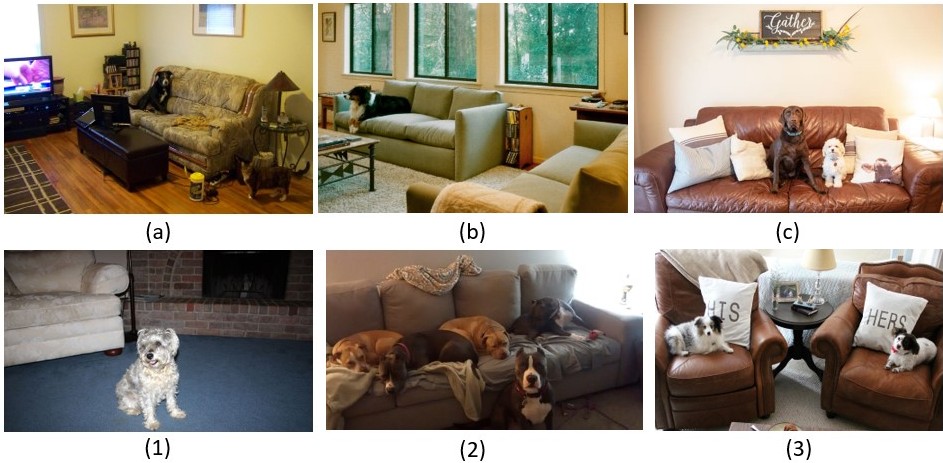}
\end{center}
    \caption{\emph{A Visual Discrimination Puzzle}}
    \label{fig:naturalvdp}

\end{figure}

In this paper, we propose a
new high-level task called a \emph{Visual Discrimination Puzzle}
(VDP). 
Figure~\ref{fig:naturalvdp} shows an example of a VDP
. 
The first row contains some
\emph{example images} $E$ ($a$, $b$, and $c$), and the second row
consists of \emph{candidate images} $C$ ($1$, $2$, and $3$). To
solve the puzzle we are asked to answer the following question:\\
\centerline{\emph{Which candidate image is most similar to all of the}}
\centerline{\emph{example images? Explain why.}}

Humans have no prior experience with VDPs, but seem to be able to
solve them with ease. We invite the reader to try solving the puzzle
in Figure~\ref{fig:naturalvdp} before reading further.


When solving a VDP, different people can (and do, in our limited experience)
come up with different answers. Many people select $\#3$ as the answer
to the puzzle in Figure~\ref{fig:naturalvdp} and explain that, in each
example image as well as candidate image $\#3$, \emph{all dogs are
  sitting on sofas} while this is not true in the other candidate
images. 

A natural formalization of the specification for VDP puzzles
is the following:
\begin{quote}
  \emph{Is there a property $P$ that holds in all example images and
    exactly one candidate image, but does not hold in the other
    candidate images?}
\end{quote}
We call such a property $P$ a \emph{discriminator}. Solving a VDP
reduces to finding a discriminator.

Observe the following salient aspects of the problem. First, the set
$E$ of example images is small. A VDP solver must learn the common
concept $P$ using only $3$-$5$ images.  Second, unlike visual question
answering (VQA), where one \emph{computes} a query over a scene,
solving VDPs requires \emph{searching} in a large space of
discriminators for one that satisfies the puzzle specification. Third,
the logical specification of the problem makes the solution sensitive
to every image. A solver cannot merely identify a candidate image that
is similar to all example images--- it must also \emph{exclude} the
other candidates. Consequently, a candidate $c$ may be a solution for one puzzle but not in another where different candidate images are present. 
For example, if we modified the \emph{non-answer} image $\#1$ in Figure~\ref{fig:naturalvdp} so that the dog \emph{is} on the sofa,
the solution might be different from $\#3$ as the concept `all dogs are on sofas' no longer corresponds to a unique candidate. Even swapping the solution candidate
image with an example image could potentially change the solution
(as there could be a \emph{simpler} discriminator for this new
puzzle that chooses a different solution). E.g., $P$ can choose $\#1$ in the first puzzle and $Q$ can be a simpler discriminator that chooses $\#2$ in the second puzzle, but is not a solution for the first puzzle.
Finally, we would accept an answer from a person or machine
only if it is justified--- we want a precise, interpretable concept
$P$ that can be evaluated on images to see that it indeed satisfies
the puzzle specification. 





\subsection{A Neurosymbolic Framework}
We propose that effective integrative frameworks can be obtained by composing learned neural models for vision and symbolic reasoning, where the latter caters to the higher-level task specification. Designing such a
framework poses several challenges, including \textbf{(1) Interface:}
determining the exact communication between the neural and symbolic
components; \textbf{(2) Interpretability:} explaining decisions of
symbolic components, potentially in terms of inputs received from
neural components; and \textbf{(3) Robustness:} symbolic components
should ideally be robust to vagaries of individual neural components
and to different implementations.

In this paper, we instantiate a neurosymbolic architecture for
solving VDPs that addresses these three challenges in novel ways.
In this work, the neural components are realized with state-of-the-art vision models 
that are trained offline on thousands of images. Given an image, the
vision model detects a \emph{scene graph} consisting of objects, their
labels, bounding boxes or relative positions, and their
relationships. Given a puzzle, we extract one scene graph for each
image in the
puzzle.  

We propose an interface to the symbolic component using
\emph{first-order logic scene models}, which can be automatically
computed from scene graphs. The symbolic component uses discrete
search (realized efficiently using a SAT solver) to synthesize a
discriminator $P$ expressible in \emph{first-order logic} over scene
models, which identifies a candidate $c$ as a solution to a given
puzzle. 
The symbolic synthesis not only solves the puzzle by finding an
appropriate candidate, but also justifies the choice using a
first-order formula that is eminently interpretable. For the puzzle in
Figure~\ref{fig:naturalvdp}, our system would potentially find the
discriminator: 
\begin{align*}
\forall x.~ (\mathsf{dog}(x) \!\implies\! ~\exists
y.~\mathsf{sofa}(y) \land \mathsf{sitting\_on}(x,y))
\end{align*}


The robustness of such a system 
certainly depends on the robustness of the vision model (for example,
it is hard to solve the puzzle in Figure~\ref{fig:naturalvdp} if a dog
is not detected).  However, there are other robustness properties of
interest. In particular, if the vision component improves and detects
new objects or relationships, we would like any discriminator found
before the improvement to \emph{remain} a discriminator with respect
to richer scene models that contain these new objects and
relationships. We call this the \emph{extension property}. We build a
domain-specific logic called \emph{First-Order Scene Logic} (FO-SL), prove that it has the extension property,
and use it to express discriminators.


The problem of synthesizing quantified discriminators in FO, and in
particular FO-SL, is a relatively new problem (as opposed to
\emph{program} synthesis). Reducing the problem to off-the-shelf
synthesizers does not scale, and we build our own SAT-based symbolic
solvers for synthesizing quantified discriminators.


%


\paragraph{Evaluation} We create three VDP datasets. Two of these are based on
real-world scenes and contain $\sim9000$ puzzles, and the other
(synthetic) dataset is based on the \clevr~ domain and consists of
{\clevrnum}
puzzles. 
We implement and evaluate our framework on the real-world datasets and
show that it is effective and robust. It solves 68\% and 80\% of the
puzzles in the two datasets and gives sensible discriminators. We
perform ablation studies that examine the effectiveness of the
domain-specific logic FO-SL
as well as the synthesis algorithm. The ablation for FO-SL involves
implementing a second synthesis solver based on a different
technique. We also compare our framework with purely neural baselines
based on image similarity models~\cite{wang2014learning} and
prototypical networks~\cite{prototype}, and we show that they perform
poorly ($\sim 40\%$). This comparison is made using the {\clevr} VDP dataset,
which is designed to have unique minimal discriminators in
FO-SL. Finally, we also create a dataset called \oddone~ consisting of
{\altnum} puzzles, which involves a different specification (picking
the ``odd one out'' from a set of images) adapted from the {\clevr}
VDP dataset. We use this dataset to evaluate how well various
approaches adapt to new high-level specifications without retraining.
%
%

\paragraph{Contributions} The primary contributions of this paper
are: (1) the VDP problem, which (cognitively) involves visual
perception as well as a search for interpretable concepts, (2) sets of
$\sim{\totalnum}$ VDP and {\oddone} puzzles that span both natural
scenes and synthetic scenes, 
(3) an instantiation of the neurosymbolic framework with a novel
interface between neural and symbolic components, the FO-SL logic for
robust visual discriminators, and an efficient discriminator synthesis
algorithm based on SAT solving, (4) an evaluation including ablation studies and comparisons to purely neural baselines.
\section{Related Work}

The idea of learning in two phases, with a first phase involving
specific concepts learned from a large dataset and a second phase
involving few-shot learning with concepts from the first phase, is not
new. The work on recognizing handwritten characters~\cite{Lake1332}
explores a similar idea. In that work,
the first phase learns a generative model of handwritten characters
using strokes and the second phase uses Bayesian learning.
In our work, we use neural models for object detection and then
SAT-based synthesis of first-order formulas.
VDP puzzles, however, are different because they require the chosen
candidate to be discriminated from other candidates, and this suggests
the use of logic. Synthesizing programs to explain behavior and
generalize has been explored in various other work
recently~\cite{graphicprog,SceneLearning19}. 

Synthesizing programs from discrete data has been studied by both the
AI and programming languages communities; the former in inductive
logic programming (ILP)~\cite{ilptext} and the latter in program
synthesis~\cite{synthesiscacm,gulwanicacm} (including the use of
SAT/SMT solvers~\cite{sketch,sygusjournal}~\hidecite{syguswebsite}).


There has been a flurry of recent work in combining neural and
symbolic learning
techniques~\cite{amizadeh2020neurosymbolic,nsvqa,Mao2019NeuroSymbolic}~\hidecite{neuralscenederendering} for problems where large datasets are available. In some cases the
goal is to learn a program (e.g., learning programs as models of
natural scenes~\cite{SceneLearning19}, assisting programmers by
learning from large code
repositories~\cite{RaychevCACM19})~\hidecite{RaychevPLDI14},
and several new techniques have
emerged~\cite{deepcoder,risingneurosymbolic,murali2018neural,robustfill,BunelICLR18,SunICML18,dilp}~\hidecite{diffprogbrock,swarathoudini}.
In this context, our work is novel in that it combines the neural and
symbolic components in two different layers, where the symbolic layer
is used to synthesize interpretable logical discriminators and handle
few-shot learning effectively.

A closely related problem is that of Visual Question Answering
(VQA)~\cite{vqa}. Neurosymbolic
approaches~\cite{nsvqa,amizadeh2020neurosymbolic} 
have been used to disentangle visual and NLP capabilities from
reasoning in order to solve VQA on artificially rendered
images.  
VQA involves \emph{queries} about a single image or scene and
end-to-end learning algorithms are commonly
used. 
However, VDP puzzles require \emph{searching} through a large class of
potential discriminators, which is inherently a higher-level task. In
a sense, we are asking whether there is \emph{some question} for which a VQA engine would say `yes' on some images and `no' on others.
discriminator. 
The work in~\cite{l3} solves few-shot classification
problems in this manner, but does not handle specifications like that of
VDPs, which involve
negation. 

The works on Neural Module Networks~\cite{nmn}~\hidecite{RonghangAutoNMN2017} and Concept Bottleneck Models~\cite{concept-bottleneck} explore methods of embedding symbolic representations or concepts into neural network architectures. However, it is unclear whether there are methods to \emph{search} for concepts (say, satisfying specifications) using these architectures.

Another related problem is solving puzzles from Raven's Progressive
Matrices~\cite{pgm,raven}. While the puzzles are similar to ours in
specification, the space of concepts is minuscule and does not require
any synthesis. The work in~\cite{raven} shows that simply enumerating
the concepts achieves 100\% accuracy on these datasets.



\section{Composing Neural Learning and Symbolic Reasoning for Discriminating Scenes}



Prior to solving VDPs, humans have first learned to distinguish
objects (\emph{dogs, sofas}, detect relationships (\emph{dog sitting
  on sofa}), poses (\emph{woman standing}), and other attributes
(\emph{cat has closed eyes}) by drawing from a rich visual experience
accumulated from childhood. Upon first encountering a VDP
puzzle, 
humans do not have the same rich experience to go by (they likely
haven't solved any VDPs previously). Despite this lack of prior
experience, they quickly formulate a mechanism to identify a solution
by \emph{composing high-level concepts that use learned lower-level
  concepts}.




\begin{figure}



\centering
\begin{center}
\includegraphics[width=0.475\textwidth]{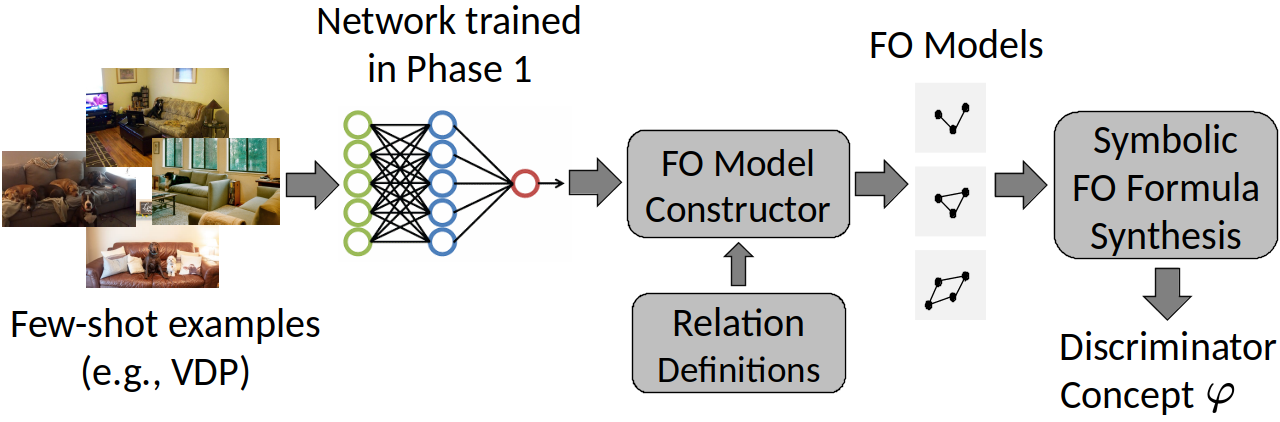}
\end{center}
%

\caption{\emph{Compositional Framework Combining Neural Learning and
    Symbolic Reasoning}}
    \label{fig:framework}
\end{figure}

Our framework is built on this intuition and involves two similar
phases. We propose a neurosymbolic approach to solve VDPs,
as illustrated in Figure~\ref{fig:framework}\footnote{A part of this image is taken from \url{https://commons.wikimedia.org/wiki/File:NeuralNetwork.png} which is public domain.}.
Phase~1 involves \emph{long-term learning} using large training sets that is
independent of any high-level task. Phase~2 depends on the
specification of a particular task (solving a
VDP), 
and it utilizes the concepts learned in Phase~1 to meet the
higher-level specification (few-shot discrimination of scenes). More
precisely, we propose:

\paragraph{Neural Learning Algorithms for Phase~1} 
Learning a scene
representation in terms of objects (``dog''), attributes (``dog is
black''), and relationships (``dog is sitting on sofa'') --- called a
scene graph --- is a well-studied problem in literature, and
convolutional neural networks (CNNs) are effective models.
In this paper we use {\sc Yolov4}~\cite{yolov4}, a CNN-based object detector
trained on the {\sc ImageNet} and {\sc COCO} datasets to predict
multiple objects with \emph{bounding boxes} and \emph{class
  labels}. 

\paragraph{Interfacing Phase~1 and Phase~2 using Scene Models}
The interface between the output of the neural component and the input
of the symbolic synthesis component is an important challenge.
We propose a novel interface, namely \emph{First-Order Scene Models},
that capture object classes, attributes, and
relationships.

\paragraph{Symbolic Synthesis Algorithms for Phase~2}
In this phase we
build novel algorithms to synthesize \emph{quantified first-order
  logic} formulas that satisfy the VDP puzzle specification and which
discriminate between the \emph{finite} first-order scene models for
each image.


If solving VDPs were the only goal, we could train models on a large
class of puzzles. Our goal instead is to focus on a different kind of
solution; we want to study how to solve puzzles and tasks afresh,
i.e., without access to a rich experience in solving them.



\section{Synthesis of First-Order Logic Discriminators}
\label{symbolic}

\subsection{First-Order Logic and Scene Models}


We work with first-order \emph{relational} logic over a signature
$\Sigma = (\mathcal{L}, \mathcal{R})$, where ${\mathcal R}$ is a
finite set of relation symbols and $\mathcal{L}\subseteq\mathcal{R}$
is a set of unary symbols we call
\emph{labels}
(which we use to model categories of objects).
Each relation symbol is associated with an
arity $n \in \mathbb{N}$, $n>0$.



The syntax of first-order logic formulas is given by:


\begin{center}
    $ \textit{Formulas~~} \varphi ::= R(x_1, \ldots, x_k) \mid x_i=x_j \mid \varphi \vee \varphi \mid \varphi \wedge \varphi \mid \neg \varphi \mid \varphi \Rightarrow \varphi \mid \exists x.~ \varphi \mid \forall x.~ \varphi $
\end{center}




We use the standard notions of models and semantics for first-order
logic (see a standard logic textbook~\cite{Enderton}).

Given a set of images $X$ for a VDP puzzle, we build a first-order
model for each image $I\in X$ by feeding $I$ to the pretrained neural
network. The model's universe corresponds to the set of objects
detected by the network. \ijcaicut{In systems such as {\sc Yolo}, these correspond to bounding boxes which have at least one label prediction that is beyond a chosen threshold.}
We model the class
labels identified by the network as unary predicates $\mathcal{L}$
(e.g. $cat(x)$ or $person(x)$), and 
the identified relationships between objects as relations
$\mathcal{R}$. The resulting \emph{First-Order Scene Models} are used
by the symbolic
synthesizer. 

\subsection{
First-Order Scene Logic and the Extension Property}

\label{sec:guardedfragment}

We argue that a first-order logic over scene models (that are obtained
from neural networks) should satisfy a particular \emph{robustness
  property}. In a suitable logic, we would like any discriminator to
remain a discriminator whenever \emph{additional} objects or
relationships are discovered by a vision model.

Suppose that we have found a discriminator 
for a puzzle, say, \emph{all dogs are on sofas}. We would like it
to remain a discriminator if new \emph{irrelevant} objects and
relationships are detected and added to the scene model. For example,
if we add to a scene model a previously unrecognized pen then our
discriminator should still work as a solution.  \ijcaicut{Perhaps
  there is an elephant in the room and our initial model doesn't
  include any elephant, but nevertheless the discriminator should
  still work when elephants are
  detected. 
  Detection of new and more fine-grained objects/relationships should
  also not make the discriminator wrong (e.g., whiskers or ears of a
  cat, small objects such as a pen on the table, etc.).}

Standard first-order logic does not have this property. For example,
in the puzzle in Figure~\ref{fig:naturalvdp}, if only dogs and sofas
were recognized then we could
express ``all dogs are on sofas'' using the formula
$\forall x. \exists y. (\textit{sofa}(y) \land (x \neq y \Rightarrow
on(x,y)))$,
which says that all objects other than a sofa are on a sofa. If the
vision model begins to recognizes tables in the images, then the
formula fails to be a discriminator (it would be false on the example
images).

 We define \emph{model extensions} as follows:

\begin{definition}[Model Extension]
A model $M'$ over
${\mathcal{R}'}$ \emph{extends} a model $M$ over $\mathcal{R}$ (where
$\mathcal{R} \subseteq \mathcal{R}'$) if $M'$ agrees with $M$ on the
interpretation of all relations in $\mathcal{R}$.
\end{definition}


Thus, given (1) the imprecision of vision
models, 
(2) 
that different systems will likely have different detection rates for
various object classes and relationships,
and (3) that the choice of visual system crucially affects the
scene model and
consequently the discriminators, we propose
the following property for
discriminators:

\begin{definition}[Extension Property]
A logic \emph{has the extension property} if any discriminator $\varphi$ for a set of models $\{M_i\}$ remains a discriminator for any extended models
$\{M'_i\}$. 
\label{def:extensionproperty}
\end{definition}

We formulate \emph{First-order Scene Logic} (FO-SL) based on guarded
logics, wherein any quantified object is always \emph{guarded} by an
assertion that it has a specific object
label:

\begin{align*}
\text{FO-SL} &::= \forall x. L(x) \Rightarrow \varphi \,\mid\,
\exists x. L(x) \wedge \varphi \,\mid\, \psi \\
\psi &::= R(\overline{x}) \,\mid\, \psi\vee\psi \,\mid\, \psi\wedge\psi
\,\mid\, \neg\psi
\,\mid\, \psi\Rightarrow\psi
\end{align*}

In the grammar above, $L$ ranges over label relations, e.g. $cat(x)$,
and $R$ ranges over relation symbols (attributes and object
relationships).

FO-SL can express properties of all cats in a scene, but not of all
objects of \emph{any} kind (a variable cannot range over cats, pens,
paintings, and specks of dust). 
We then show the following result (see Appendix~\ref{app:extension-property-proof} for the proof):


\begin{theorem}
The guarded fragment has the extension property.
\end{theorem}
We thus propose the use of FO-SL as our space of possible
discriminators.

\subsection{
  Solving VDP 
  by Synthesizing Formulas}
 \label{subsec:formulation}

 We solve VDPs by synthesizing formulas that serve as discriminators
 for scene models. Let us fix a puzzle with example images $E$ and
 candidate images $C$, as well as a signature $\Sigma$ that is
 determined by a given vision model. Let the corresponding scene
 models be
$E_M = \{ e_1^M, \ldots, e_u^M \}$ and $C_M = \{ c_1^M, \ldots, c_v^M \}$.



\begin{definition}[ Discriminator]
An first-order sentence $\varphi$ is a \emph{discriminator} for
$(E_M, C_M)$ if there is a model $\widehat{c}^M \in C_M$ such that:

\noindent
(D1) For every model $e^M \in E_M$,
               $e^M \models \varphi$

\noindent
(D2) $\widehat{c}^M \models \varphi$

\noindent
(D3) For every $c^M \in C_M$
           such that $c^M \not = \widehat{c}^M$,
            $c^M \not \models \varphi$

\end{definition}
This definition formally captures the puzzle specification: a
discriminator is a formula that is true in all example images (D1) and
true in \emph{exactly} one candidate image (D2 and D3).

\subsubsection{Learning FO Discriminators}
We describe an algorithm for synthesizing FO-SL discriminators to solve a given VDP puzzle. In
particular, we build an algorithm that finds \emph{conjunctive}
discriminators. 

We tried using state-of-the-art \emph{program synthesis} engines that
handle the
{\sc SyGuS} format (Syntax-Guided
Synthesis)~\cite{sygusjournal}~\hidecite{syguswebsite}. These
did not scale well 
(see Section~\ref{sec:rqs}), so we implemented our own solution using
SAT solvers.
%


If $k$ is the number of quantifiers in the discriminator, then we
initially set $k$ to $1$ and iteratively increment it whenever we
cannot find a discriminator with $k$ quantifiers. We introduce a
Boolean variable $b_a$ for every atomic formula $a(\overline{x})$ that
can occur in the matrix, i.e., the quantifier-free part of the
formula. The intention is that $b_a$ is true if and only if the atomic
formula $a$ occurs as a conjunct in the matrix of the
discriminator. We also introduce $k$ Boolean variables that determine
whether the $k$ quantifiers are existential or universal, as well as
variables that determine the guards (labels in FO-SL) for each
quantified variable. Given a valuation $\overline{b}$ for these
variables, we can write a formula $\phi(\overline{b})$ that evaluates
the discriminator encoded by $\overline{b}$ on all scene models.
With extra Boolean variables that encode the choice of candidate scene
model, 
we can formulate a constraint that reflects the specification of the
puzzle and the definition of discriminator above.

We then use a SAT solver (such as Z3~\cite{Z3}) to determine whether
the constraint is satisfiable.
If the SAT solver finds a satisfying valuation, we can construct the
FO-SL discriminator and the chosen candidate from the valuation. If
the constraint is unsatisfiable, then we know there is no conjunctive
discriminator with $k$ quantifiers.

\section{Evaluation}
\seclabel{evaluation}

\subsection{Datasets}

We create 11,600 puzzles across four datasets\ijcaicut{: Natural
  Scenes and GQA VDP datasets over real-world scenes,
  {\clevr} VDP dataset over {\clevr} domain
  scenes~\cite{clevrdataset}, and an \oddone~ puzzles dataset based on
  {\clevr} scenes}.  We describe these briefly. We also invite the reader to browse the static website of VDPs\footnote{The datasets, code, and the website of VDPs can be found at: \textcolor{blue}{https://github.com/muraliadithya/vdp}} for a sample of
puzzles across the datasets.
\paragraph{Natural Scenes} The Natural Scenes VDP dataset is created
from \natbase~\emph{base} real-world concept classes such as ``all
dogs are on sofas''. For each 
class, we collect positive images that satisfy the concept and
negative images that do not. \ijcaicut{(but satisfy a variation of it,
  e.g, `Some dogs on sofas'). These images were largely chosen from
  the internet and public domain datasets
  (see Appendix for attributions).} We create puzzles by choosing all
examples from the positive set and all candidates from the negative
set except for
one positive image (the \emph{intended} candidate).
We sample {\natnum} puzzles
randomly. We provide a description of these concepts in Table
\ref{yolotable}.

\begin{table}
  \begin{tabular}{@{}rlrl@{}}
    \toprule
    \textbf{ID} & \textbf{Class Description} &  \textbf{ID} & \textbf{Class Description} \\
    \midrule

      1 & All teddy bears on a                & 2 & There is an SUV \\
      & sofa & & \\
      3 & Onward lane (person               & 4 & Fruit in separate piles \\
        & on left sidewalk)                     &   &  (apples \& oranges)   \\
      5 & Kickoff position (ball                 & 6 & Laid out place setting \\
         & b/w two people)                &   &  (v/s dirty dishes) \\
      7 & Person kicking ball                      & 8 & Dog herding sheep \\
      9 & Parking spot                             & 10 & People carrying \\
        &                                          &    & umbrellas \\
      11 & Bus filled with people                   & 12 & All dogs on sofas \\
      13 & Desktop PC                               & 14 & People wearing ties \\
      15 & Person sleeping on                 & 16 & All cats on sofas \\
      & bench &&\\
      17 & Kitchen                                  & 18 & TV is switched on \\
      19 & Two cats on same sofa                    & 20 & Cat displayed on TV \\

    \bottomrule
  \end{tabular}

     \caption{Concept Class Descriptions for Natural Scenes Dataset}

      \label{yolotable}
\end{table}


\begin{table}
\centering
\begin{tabular} 
{@{}l@{\extracolsep{\fill}}l@{}}
\toprule
 \textbf{ID} & \textbf{Concept Class Schema} \\
\midrule
1  \quad & Every \texttt{shapeX} has a \texttt{shapeY} to  its left and right.                    \\
2  \quad & There is a \texttt{shapeX} of color \texttt{colorA} to the left of a \\& \texttt{shapeY} of color \texttt{colorB}.           \\
3  \quad & There is a \texttt{shapeX} to the right of every  \texttt{shapeY}.                                   \\
4  \quad & There is a \texttt{shapeX} and there is a  \texttt{shapeY} to the left \\& of all \texttt{shapeX}.                                     \\
5  \quad & Every \texttt{shapeX} has a \texttt{shapeY} to its right.                                                          \\
6  \quad & All \texttt{shapeX}s and \texttt{shapeY}s have the same color.                                                         \\
7  \quad & There is no color such that there is only  one \texttt{shapeX} \\& of that color.                                  \\
8  \quad & There is a leftmost \texttt{shapeX} and a rightmost  \texttt{shapeX}.                                                     \\
9  \quad & All \texttt{shapeX} are to the left of all \texttt{shapeY}s  and the \\&  rightmost \texttt{shapeY} is  made of \texttt{materialQ}.   \\
10 ~\quad & All \texttt{shapeX}s are to the left of a  \texttt{shapeY} of 
color \\&\texttt{colorB}.                                                     \\
11 ~\quad & All \texttt{shapeX}s are to the left of all \texttt{shapeY}s.                                                          \\
12 ~\quad & Every \texttt{shapeX} has a \texttt{shapeY} to its right.                                                             \\
13 ~\quad & Every \texttt{shapeX} has a \texttt{shapeY} behind it.                                                                \\
14 ~\quad & There are three \texttt{shapeX}s of the same color.                                                        \\
15 ~\quad & There is a \texttt{materialQ} \texttt{shapeX} to the left  of    \\&all \texttt{shapeY}s.                                                \\

\bottomrule
\end{tabular}
\caption{ Concept Class Schema for {\clevr} VDP Dataset. \texttt{shapeX} and \texttt{shapeY} (disequal) range over sphere, cylinder, and cube. \texttt{colorA} and \texttt{colorB} (disequal) range over 8 possible colors. \texttt{materialQ} can either be rubber or metal.}

\label{clevrtable}

\end{table}

\paragraph{GQA VDP dataset} The GQA VDP dataset is created automatically
using the GQA dataset~\cite{gqa}, which is a
VQA dataset. It consists of real-world 
scenes along with scene graphs, as well as questions and answers about the images. \ijcaicut{ The GQA dataset was collected using human annotators.} 
We use questions with yes/no answers such as: \emph{Is there a fence
  made of wood?} We create puzzles as described above (with ``yes''
images being the positive category) and sample {\gqanum} random
VDPs. Note that the proposition corresponding to the question,
i.e. \emph{there is a fence made of wood}, is hence a discriminator.

\paragraph{{\bfclevr} VDP dataset.} 
The {\clevr}
domain~\cite{clevrdataset} is an artificial VQA domain consisting of
images with 3D shapes
such as spheres, cubes, etc.
The objects possess a rich combination of attributes such as
shape, colour, size, and material. 
The {\clevr} VDP dataset consists of {\clevrbase} base concept classes
as described in Table~\ref{clevrtable}. Each concept class is a
\emph{schema} such as `ShapeX and ShapeY have the same color', where
the variables ShapeX and ShapeY denote distinct shapes. We create
abstract VDPs whose unique minimal discriminator is contained in the
FO-SL schema (e.g.,
$\forall x. \mathit{ShapeX} \Rightarrow \forall y. \mathit{ShapeY}
\Rightarrow \mathit{samecolor}(x,y)$). We instantiate these variables
with various combinations of 
shapes, colours, etc., 
and we sample \ijcaicut{\clevrvariants~ \emph{variant} puzzles per
  concept schema. We also perform some swaps of images to create some
  variants.  This yields} {\clevrnum} VDPs with unique solutions in
the FO-SL conjunctive
fragment. 
See Appendix~\ref{app:clevrdataset} for more details.

\paragraph{{\bfoddone} Puzzles dataset.} We create a
dataset of 
discrimination puzzles that are different from VDPs. An \oddone~
puzzle consists of $4$ images
with the objective of identifying the image that is the \emph{odd one
  out}. 
We formalize this task similar to VDPs by demanding a concept
$\varphi$ that satisfies \emph{exactly} three
images. 
We create these
from the {\clevr} VDP dataset by choosing three example images and one
non-answer candidate image from each
puzzle. 
Finally, we only include the {\altnum} puzzles  
that have a unique minimal discriminator in FO-SL. 

\noindent
Note that only the {\clevr} VDP and {\oddone} datasets have unique
discriminators in conjunctive FO-SL.

\subsection{Implementation}
\label{sec:implementation}

We use a pretrained model of {\sc Yolo}v4~\cite{yolov4}
for 
the Natural Scenes dataset, 
which outputs object labels and bounding boxes. 
For the \clevr~ domain, we use a pretrained model from the work in~\cite{nsvqa} \ijcaicut{whose architecture is a Mask-RCNN followed by a ResNet-34\cite{he2016deep}}. The model produces a set of objects with
their coordinates in 3D space and other attributes like shape, color,
or material. We compute first-order scene models based on the outputs
automatically.

The symbolic synthesizer implements an algorithm that searches for
conjunctive discriminators in FO-SL. The algorithm creates the
constraints 
described in 
Section~\ref{subsec:formulation} and uses the SAT-solver {\sc
  Z3}~\cite{Z3}. 
We also implement 
another synthesizer that searches for discriminators in full
first-order logic\ijcaicut{, including non-conjunctive concepts and
  arbitrary Boolean combinations. We use this solver} to perform
ablation studies. 
See Appendix~\ref{app:implementation} for more details about implementation.

\subsection{Experiments}
\label{sec:rqs}
We report on the evaluation of our tool on the datasets and various
ablation studies in terms of the research questions (RQs) below.

\subsubsection*{Experiments on Natural Scenes and GQA VDP Datasets}

\subsubsection{RQ1: Effectiveness}
We evaluate our tool on the Natural
Scenes dataset, applying the SAT-based symbolic synthesizer with a
timeout. Since it is always possible to discriminate a finite set of
(distinct) models with a complex (but perhaps unnatural) formula, we
restrict our search to discriminators with complexity smaller than or
equal to the target discriminator. 
We present the results in Figure~\ref{fig:natsceneseval}. Our tool is
effective and solves 68\% of the {\natnum} puzzles, with an average
accuracy of 71\% per concept class.

\begin{figure}

    \begin{center}
            \scalebox{0.42}{\input{stackedbarchartsmall.pgf}}
    \end{center}
        \caption{Evaluation on {\natnum} Natural Scenes Dataset
          puzzles: Class ID refers to the ID in Table~\ref{yolotable},
          and (N) is the number of puzzles in each
          class. Intended: solver chose the intended
          discriminator. Unintended: solver did not choose
          the intended discriminator.}
        \label{fig:natsceneseval}

\end{figure}

Our tool finds varied discriminators with multiple quantifiers and
complex nesting. For example, ``Football in between two people''
(kickoff position) is expressed by
$\forall x.\,(\mathsf{person}(x)\Rightarrow~ \forall
y.\,(\mathsf{sports~ball}(y)\Rightarrow ~\exists
z.\,(\mathsf{person}(z) \wedge \mathsf{left}(y,x) \land
\mathsf{right}(y,z))))$, which has three quantifiers with
alternation. The tool also generalizes intended
discriminators.

Note that the Natural Scenes VDPs do not have unique solutions since
smaller discriminators may
exist. 
For example, in the concept class of ``Laid out place settings'', a
particular puzzle always displayed cups in example images. Our tool
picked an unintended candidate satisfying ``There is a cup'' rather
than the intended candidate satisfying a more complex discriminator
(utensil
arrangements). 
Our tool picks an unintended candidate in 17\% of solutions. We
provide more examples of discriminators found in Table~\ref{largeyolotable} (Appendix).

\begin{figure}
    \begin{center}
            \scalebox{0.42}{\input{barplot.pgf}}
    \end{center}
        \caption{Comparison with neural baselines on \clevrnum~ {\clevr} VDP Dataset puzzles: Class ID refers to the ID in Table~\ref{clevrtable}, and (N) is the number of puzzles in each class. \textbf{Triplet}: image similarity baseline. \textbf{Prototype}: prototypical network baseline. \textbf{Ours}: our solver. Dashed line represents accuracy of a random predictor. }
        \label{fig:similaritybaseline}
\end{figure}

\subsubsection{RQ2: Generality} 
How general
is our framework? The GQA VDP dataset consists of automatically
generated puzzles where neither the images nor the discriminators are
curated. However, the pipeline fails if the vision model fails. In
fact, most failures on the Natural Scenes dataset are of this kind. In
our experience, although vision models are good at detecting objects,
many report several bounding boxes for the same object and are not
accurate. We experimented with SOTA scene graph
generators (similar to the work in~\cite{rowan}) and
found many issues.

To better study the generality of our formulation, we ablate the
vision model errors and use the ground-truth scene graphs of images
given in the GQA dataset to extract scene models. Our tool performs
very well: it solves 81\% of the puzzles (4\% unintended). See the website of VDPs for examples of (un)solved puzzles. 

\subsubsection{RQ3: Failure Modes}
The primary failure mode of the tool is
failure of the vision model, as discussed above. This failure
manifests as nonsensical discriminators. We leave the problem of
designing frameworks that are robust despite vision model failures to
future
work. 

We identify two other failure modes where the solver was not able to
find any discriminator, namely (1) expressive power of the interface:
we do not solve puzzles with discriminators like \emph{There is a bag
  in the bottom portion of the image}, since information about the
region of the image in which an object was present is not typically
expressed in scene graphs. Many of the unsolved puzzles in the GQA VDP
dataset belong to this failure mode; and (2) expressive power of the
FO-SL fragment: consider
the 
concept class \emph{TV is switched on} in the Natural Scenes dataset,
expressed as 
$\exists x.\,\mathsf{tv}(x)\land \exists y.\,\mathsf{displayed~on}(y,
x)$. This requires unguarded quantification and cannot be expressed in
FO-SL.  
Another example is \emph{There is a dog that is not white}, which
requires negation and is not expressible in the conjunctive fragment.

\subsubsection{RQ4: Ablation of Synthesis Algorithm}
Our SAT-based synthesis algorithm 
is quite effective and finds discriminators in a few
seconds.  
However, there are other possible synthesis engines, in particular
those developed in the program synthesis literature
for 
SyGuS
problems~\cite{sygusjournal}~\hidecite{syguswebsite}. 
The specification of a discriminator can be
encoded 
as a SyGuS problem, and so we perform an ablation study using {\sc
  CVC4Sy}~\cite{cvc4sy} (winner of SyGuS competitions) to find
discriminators. {\sc CVC4Sy} did not scale and took at least 10
minutes for even moderately difficult puzzles, often not terminating
after 30 minutes.

\subsubsection{RQ5: Ablation of FO-SL}
Evaluation of RQ1 shows that FO-SL is
a rich logic that expresses 
many interesting discriminators.
However, 
would a more expressive logic find more natural or better
discriminators? 
We perform an ablation study on Natural Scenes VDPs by implementing a
second solver for symbolic synthesis. This solver searches for
discriminators in full first-order logic rather than
FO-SL: 
the quantifiers are not guarded and the
matrix 
need not be conjunctive. Note that FOL does not satisfy the extension
property.


This solver is slower and times out for 57\% of the puzzles. Among
solved puzzles, solutions are sometimes more
general, 
e.g. \emph{There is something that is within all sofas}, instead of
\emph{All dogs on sofas}. However,
we almost
always obtain unnatural discriminators,
e.g. \emph{There is no chair and there is some
  non-sofa}.  
We therefore conclude that the guarded quantification in FO-SL is
important, and full first-order logic is not 
suitable for finding natural discriminators. See Appendix~\ref{app:bottomupsolver} for a detailed description of the solver as well as Table~\ref{nonguardedtablelarge} in the Appendix for more
examples of unnatural discriminators.

\subsubsection*{Experiments on {\clevr} VDP Dataset}

\subsubsection{RQ4: Neural Baselines}
The {\clevr} VDP dataset consists of
puzzles with unique discriminators by construction, and our tool
performs (unsurprisingly) very well (99\%). 
Since solutions are unique, we can ask if solving VDPs is learnable
using neural models.

We first construct a baseline from an image similarity model trained
using triplet loss~\cite{wang2014learning}. To solve a VDP, we choose
the candidate that maximizes the product of similarity scores with all
the example images. 
We present the class-wise performance of this model in
Figure~\ref{fig:similaritybaseline}. It does not perform very well and
is rarely better than chance (33\%).
This shows that VDP solutions are not aligned well with commonly
learned image features. 
Therefore, we evaluate another baseline based on prototypical
networks~\cite{prototype} by training on VDPs. Such networks aim to
learn embeddings such that the average embedding of a class of images
acts as a prototype, which can then be used to compute similarity with
respect to the
class. 
We fine-tune a ResNet18~\cite{he2016deep} + MLP architecture
pretrained on CIFAR10 using $6$ concept classes, validated against a
held-out set of classes. We then evaluate it on the unseen classes and
other unseen puzzles.  
This model achieves a slightly better overall accuracy of
40\% 
as shown in Figure~\ref{fig:similaritybaseline}, but is still not
performant.

\subsubsection*{Experiments on \bfoddone~ Puzzles Dataset}

\subsubsection{RQ7: Adaptive Mechanisms}
We evaluate the adaptability of
mechanisms (without retraining) when the higher-level puzzle
description is changed. Our framework is evidently highly adaptable
and we can solve \oddone~ puzzles without retraining by simply
changing the synthesis objective (the
constraint).

We evaluate the adaptability of the baseline models learned for VDP on
the new task by adapting the scoring function (see Appendix~\ref{app:oddonedatasetevaluation}) and find
that they perform very poorly. The similarity and prototypical network
baselines perform at 13\% and 23\% respectively, compared to a random
predictor at
25\%. 
Therefore, we conclude that the neural representations learned using
the two baselines for one task do not lend themselves well to newer
high-level task specifications without retraining.

\section{Conclusions}
The results of this paper argue that building symbolic synthesis and reasoning over outputs of a neural network lead to robust interpretable reasoning for solving puzzles such as VDP. For a future direction, the following problem formulation for VDPs seems interesting: given a VQA engine, find a property expressed in natural language that acts as a discriminator, as interpreted by the VQA engine.
Solving the above would result in natural language discriminators (which would be more general than FO-SL), but the problem of searching over such discriminators seems nontrivial. 
We would also like to adapt the neurosymbolic approach in this paper to real-world applications that require learning interpretable logical concepts from little data, including obtaining differential diagnoses from medical images and health records, and learning behavioral rules for robots from interaction with humans and the environment.
\section*{Acknowledgments}
This work is  supported in part by a research grant from Amazon and a Discovery Partner’s Institute (DPI) science team seed grant.

{\RaggedRight
\bibliographystyle{named}
\bibliography{refs}

\begin{thebibliography}{}

\bibitem[\protect\citeauthoryear{Alur \bgroup \em et al.\egroup
  }{2015}]{sygusjournal}
Rajeev Alur, Rastislav Bod{\'\i}k, Eric Dallal, Dana Fisman, Pranav Garg,
  Garvit Juniwal, Hadas Kress-Gazit, P.~Madhusudan, Milo M.~K. Martin, Mukund
  Raghothaman, Shambwaditya Saha, Sanjit~A. Seshia, Rishabh Singh, Armando
  Solar-Lezama, Emina Torlak, and Abhishek Udupa.
\newblock Syntax-guided synthesis.
\newblock {\em Dependable Software Systems Engineering}, 2015.

\bibitem[\protect\citeauthoryear{Alur \bgroup \em et al.\egroup
  }{2018}]{synthesiscacm}
Rajeev Alur, Rishabh Singh, Dana Fisman, and Armando Solar-Lezama.
\newblock Search-based program synthesis.
\newblock {\em Commun. ACM}, 61(12):84--93, November 2018.

\bibitem[\protect\citeauthoryear{Amizadeh \bgroup \em et al.\egroup
  }{2020}]{amizadeh2020neurosymbolic}
Saeed Amizadeh, Hamid Palangi, Oleksandr Polozov, Yichen Huang, and Kazuhito
  Koishida.
\newblock Neuro-symbolic visual reasoning: Disentangling "visual" from
  "reasoning".
\newblock {\em ICML}, 2020.

\bibitem[\protect\citeauthoryear{Andreas \bgroup \em et al.\egroup
  }{2016}]{nmn}
Jacob Andreas, Marcus Rohrbach, Trevor Darrell, and Dan Klein.
\newblock Neural module networks.
\newblock {\em CVPR}, 2016.

\bibitem[\protect\citeauthoryear{Andreas \bgroup \em et al.\egroup }{2018}]{l3}
Jacob Andreas, Dan Klein, and Sergey Levine.
\newblock Learning with latent language.
\newblock {\em ACL}, 2018.

\bibitem[\protect\citeauthoryear{Antol \bgroup \em et al.\egroup }{2015}]{vqa}
Stanislaw Antol, Aishwarya Agrawal, Jiasen Lu, Margaret Mitchell, Dhruv Batra,
  C.~Lawrence Zitnick, and Devi Parikh.
\newblock Vqa: Visual question answering.
\newblock {\em ICCV}, 2015.

\bibitem[\protect\citeauthoryear{Balog \bgroup \em et al.\egroup
  }{2017}]{deepcoder}
Matej Balog, Alexander~L. Gaunt, Marc Brockschmidt, Sebastian Nowozin, and
  Daniel Tarlow.
\newblock Deepcoder: Learning to write programs.
\newblock {\em ICLR}, 2017.

\bibitem[\protect\citeauthoryear{Bunel \bgroup \em et al.\egroup
  }{2018}]{BunelICLR18}
Rudy Bunel, Matthew~J. Hausknecht, Jacob Devlin, Rishabh Singh, and Pushmeet
  Kohli.
\newblock Leveraging grammar and reinforcement learning for neural program
  synthesis.
\newblock {\em {ICLR}}, 2018.

\bibitem[\protect\citeauthoryear{de Moura and Bj{\o}rner}{2008}]{Z3}
Leonardo de~Moura and Nikolaj Bj{\o}rner.
\newblock Z3: An efficient smt solver.
\newblock {\em TACAS}, 2008.

\bibitem[\protect\citeauthoryear{Devlin \bgroup \em et al.\egroup
  }{2017}]{robustfill}
Jacob Devlin, Jonathan Uesato, Surya Bhupatiraju, Rishabh Singh, Abdel-rahman
  Mohamed, and Pushmeet Kohli.
\newblock Robustfill: Neural program learning under noisy i/o.
\newblock {\em ICML}, 2017.

\bibitem[\protect\citeauthoryear{Ellis \bgroup \em et al.\egroup
  }{2018}]{graphicprog}
Kevin Ellis, Daniel Ritchie, Armando Solar-Lezama, and Joshua~B. Tenenbaum.
\newblock Learning to infer graphics programs from hand-drawn images.
\newblock {\em NIPS}, 2018.

\bibitem[\protect\citeauthoryear{Enderton}{2011}]{Enderton}
Herbert~B. Enderton.
\newblock {\em A Mathematical Introduction to Logic}.
\newblock Academic Press, 2011.

\bibitem[\protect\citeauthoryear{Evans and Grefenstette}{2018}]{dilp}
Richard Evans and Edward Grefenstette.
\newblock Learning explanatory rules from noisy data.
\newblock {\em J. Artif. Int. Res.}, 61(1):1–64, Jan 2018.

\bibitem[\protect\citeauthoryear{Goodfellow \bgroup \em et al.\egroup
  }{2016}]{Goodfellow16}
Ian Goodfellow, Yoshua Bengio, and Aaron Courville.
\newblock {\em Deep Learning}.
\newblock MIT Press, 2016.
\newblock \url{http://www.deeplearningbook.org}.

\bibitem[\protect\citeauthoryear{Gulwani \bgroup \em et al.\egroup
  }{2015}]{gulwanicacm}
Sumit Gulwani, Jos{\'e} Hern\'{a}ndez-Orallo, Emanuel Kitzelmann, Stephen~H.
  Muggleton, Ute Schmid, and Benjamin Zorn.
\newblock Inductive programming meets the real world.
\newblock {\em Commun. ACM}, 58(11):90--99, October 2015.

\bibitem[\protect\citeauthoryear{He \bgroup \em et al.\egroup
  }{2016}]{he2016deep}
Kaiming He, Xiangyu Zhang, Shaoqing Ren, and Jian Sun.
\newblock Deep residual learning for image recognition.
\newblock {\em CVPR}, 2016.

\bibitem[\protect\citeauthoryear{Hudson and Manning}{2019}]{gqa}
Drew~A. Hudson and Christopher~D. Manning.
\newblock Gqa: A new dataset for real-world visual reasoning and compositional
  question answering.
\newblock {\em CVPR}, 2019.

\bibitem[\protect\citeauthoryear{Johnson \bgroup \em et al.\egroup
  }{2017}]{clevrdataset}
J.~Johnson, B.~Hariharan, Laurens van~der Maaten, Li~Fei-Fei, C.~L. Zitnick,
  and Ross~B. Girshick.
\newblock Clevr: A diagnostic dataset for compositional language and elementary
  visual reasoning.
\newblock {\em CVPR}, 2017.

\bibitem[\protect\citeauthoryear{Koh \bgroup \em et al.\egroup
  }{2020}]{concept-bottleneck}
Pang~Wei Koh, Thao Nguyen, Yew~Siang Tang, Stephen Mussmann, Emma Pierson, Been
  Kim, and Percy Liang.
\newblock Concept bottleneck models.
\newblock In {\em ICML}, 2020.

\bibitem[\protect\citeauthoryear{Lake \bgroup \em et al.\egroup
  }{2015}]{Lake1332}
Brenden~M. Lake, Ruslan Salakhutdinov, and Joshua~B. Tenenbaum.
\newblock Human-level concept learning through probabilistic program induction.
\newblock {\em Science}, 350(6266):1332--1338, 2015.

\bibitem[\protect\citeauthoryear{Liu \bgroup \em et al.\egroup
  }{2019}]{SceneLearning19}
Yunchao Liu, Jiajun Wu, Zheng Wu, Daniel Ritchie, William~T. Freeman, and
  Joshua~B. Tenenbaum.
\newblock Learning to describe scenes with programs.
\newblock {\em ICLR}, 2019.

\bibitem[\protect\citeauthoryear{Mao \bgroup \em et al.\egroup
  }{2019}]{Mao2019NeuroSymbolic}
Jiayuan Mao, Chuang Gan, Pushmeet Kohli, Joshua~B. Tenenbaum, and Jiajun Wu.
\newblock {The Neuro-Symbolic Concept Learner: Interpreting Scenes, Words, and
  Sentences From Natural Supervision}.
\newblock {\em ICLR}, 2019.

\bibitem[\protect\citeauthoryear{Murali \bgroup \em et al.\egroup
  }{2018}]{murali2018neural}
Vijayaraghavan Murali, Letao Qi, Swarat Chaudhuri, and Chris Jermaine.
\newblock Neural sketch learning for conditional program generation.
\newblock {\em ICLR}, 2018.

\bibitem[\protect\citeauthoryear{N\'{e}dellec}{1998}]{ilptext}
Claire N\'{e}dellec.
\newblock Inductive logic programming, from machine learning to software
  engineering by f. bergadano and d. gunetti, the mit press, usa, 1996.
\newblock {\em Knowl. Eng. Rev.}, 13(2):201–208, jul 1998.

\bibitem[\protect\citeauthoryear{Parisotto \bgroup \em et al.\egroup
  }{2017}]{risingneurosymbolic}
Emilio Parisotto, Abdel{-}rahman Mohamed, Rishabh Singh, Lihong Li, Dengyong
  Zhou, and Pushmeet Kohli.
\newblock Neuro-symbolic program synthesis.
\newblock {\em ICLR}, 2017.

\bibitem[\protect\citeauthoryear{Raychev \bgroup \em et al.\egroup
  }{2019}]{RaychevCACM19}
Veselin Raychev, Martin~T. Vechev, and Andreas Krause.
\newblock Predicting program properties from 'big code'.
\newblock {\em Commun. {ACM}}, 62(3):99--107, 2019.

\bibitem[\protect\citeauthoryear{Reynolds \bgroup \em et al.\egroup
  }{2019}]{cvc4sy}
Andrew Reynolds, Haniel Barbosa, Andres N{\"o}tzli, Clark Barrett, and Cesare
  Tinelli.
\newblock cvc4sy: Smart and fast term enumeration for syntax-guided synthesis.
\newblock {\em CAV}, 2019.

\bibitem[\protect\citeauthoryear{Santoro \bgroup \em et al.\egroup
  }{2018}]{pgm}
Adam Santoro, Felix Hill, David G.~T. Barrett, Ari~S. Morcos, and Timothy~P.
  Lillicrap.
\newblock Measuring abstract reasoning in neural networks.
\newblock {\em ICML}, 2018.

\bibitem[\protect\citeauthoryear{Snell \bgroup \em et al.\egroup
  }{2017}]{prototype}
Jake Snell, Kevin Swersky, and Richard Zemel.
\newblock Prototypical networks for few-shot learning.
\newblock {\em NIPS}, 2017.

\bibitem[\protect\citeauthoryear{Solar-Lezama \bgroup \em et al.\egroup
  }{2006}]{sketch}
Armando Solar-Lezama, Liviu Tancau, Rastislav Bodik, Sanjit Seshia, and Vijay
  Saraswat.
\newblock Combinatorial sketching for finite programs.
\newblock {\em SIGOPS Oper. Syst. Rev.}, 2006.

\bibitem[\protect\citeauthoryear{Sun \bgroup \em et al.\egroup
  }{2018}]{SunICML18}
Shao{-}Hua Sun, Hyeonwoo Noh, Sriram Somasundaram, and Joseph Lim.
\newblock Neural program synthesis from diverse demonstration videos.
\newblock {\em {ICML}}, 2018.

\bibitem[\protect\citeauthoryear{Wang \bgroup \em et al.\egroup
  }{2014}]{wang2014learning}
Jiang Wang, Yang Song, Thomas Leung, Chuck Rosenberg, Jingbin Wang, James
  Philbin, Bo~Chen, and Ying Wu.
\newblock Learning fine-grained image similarity with deep ranking.
\newblock {\em CVPR}, 2014.

\bibitem[\protect\citeauthoryear{Wang \bgroup \em et al.\egroup
  }{2021}]{yolov4}
Chien-Yao Wang, Alexey Bochkovskiy, and Hong-Yuan~Mark Liao.
\newblock Scaled-yolov4: Scaling cross stage partial network.
\newblock {\em CVPR}, 2021.

\bibitem[\protect\citeauthoryear{Yi \bgroup \em et al.\egroup }{2018}]{nsvqa}
Kexin Yi, Jiajun Wu, Chuang Gan, Antonio Torralba, Pushmeet Kohli, and
  Joshua~B. Tenenbaum.
\newblock Neural-symbolic vqa: Disentangling reasoning from vision and language
  understanding.
\newblock {\em NeurIPS}, 2018.

\bibitem[\protect\citeauthoryear{Zellers \bgroup \em et al.\egroup
  }{2018}]{rowan}
Rowan Zellers, Mark Yatskar, Sam Thomson, and Yejin Choi.
\newblock Neural motifs: Scene graph parsing with global context.
\newblock {\em CVPR}, 2018.

\bibitem[\protect\citeauthoryear{Zhang \bgroup \em et al.\egroup
  }{2019}]{raven}
Chi Zhang, Feng Gao, Baoxiong Jia, Yixin Zhu, and Song-Chun Zhu.
\newblock Raven: A dataset for relational and analogical visual reasoning.
\newblock {\em CVPR}, 2019.

\end{thebibliography}
}

\appendix
\section*{General Information}


\paragraph{Citations for assets: }
Tools used for the pipeline are presented below (and also on the website).
\begin{itemize}
        \item \textbf{CLEVR dataset generator}:The image generation script was modified and used for generating the CLEVR domain VDP puzzles. Original code available at: \textcolor{blue}{https://github.com/facebookresearch/clevr-dataset-gen}. Modified code available at:  \textcolor{blue}{https://github.com/anonymousocean/vdp-tool-chain/tree/master/clevr-generate}. The original code is available under a BSD license. 
        \item \textbf{Natural Scenes inference engine}: Object recognition in the natural scenes dataset was performed using YOLOv4. The original weights and code were used with slight modifications. Original code available at:  \textcolor{blue}{https://github.com/pjreddie/darknet}. Modified code available at:  \textcolor{blue}{https://github.com/anonymousocean/vdp-tool-chain/tree/master/vdp/darknet}. The original code is available under a MIT license. 
        \item \textbf{NS-VQA inference pipeline}: The MaskRCNN + Attribute net model was used for inferring object attributes from the CLEVR domain puzzles. Original code available at:  \textcolor{blue}{https://github.com/kexinyi/ns-vqa}. Modified code available at:  \textcolor{blue}{https://github.com/anonymousocean/vdp-tool-chain/tree/master/clevr-inference}. The original code is available under a MIT license. 
        \item \textbf{Deep Ranking Similarity Baseline}: The deepranking model and code was used as a baseline for our tool.  Original code available at:  \textcolor{blue}{https://github.com/USCDataScience/Image-Similarity-Deep-Ranking}. Modified code available at:  \textcolor{blue}{https://github.com/anonymousocean/vdp-tool-chain/tree/master/pt-baseline}. The original code is available under an Apache v2 license. 

        
        \item \textbf{Images for Natural Scenes dataset}: Since these images are taken from the internet, we can only allow access under fair use for research if users sign a permission form. Please contact us to get access.
\end{itemize}

\paragraph{Total amount of compute and the resources used}: We used an AWS \texttt{p2.xlarge} (4vCPUs, 61GB RAM, 150GB HDD, 1 NVIDIA K80 GPU) instance for running our experiments. Apart from time taken to generate the various datasets, the experiments took about 120 hours to attempt solving all 11000 puzzles.




\section{Appendix}
\label{app}

\subsection{Extension Property of FO-SL}
\label{app:extension-property-proof}
\begin{theorem*}
  The guarded fragment (i.e., FO-SL) has the extension property.
\end{theorem*}
\begin{proof*}
  Consider a set of models $\{M_i\}$ and a discriminator $\varphi$ for
  that set. Let $\{M'_i\}$ be a set of models that extends $\{M_i\}$
 with respect to $\varphi$. Observe that $\varphi$ is a sentence of
  the form $Q_1{\,:\,}x_1\cdots Q_n{\,:\,}x_n \psi(\vec{x})$, where $\psi$ has no
  quantifiers, and each $Q_i{\,:\,}L(x_i)$ is shorthand for a guarded
  quantifier with label $L$. The truth of the sentence $\varphi$ is
  completely determined by how $\psi(\vec{x})$ evaluates for those
  assignments to $\vec{x}$ that make all the guards $L_i$ true. Since
  each $M'_i$ (over some $\mathcal{R'}$) extends $M_i$ over $\mathcal{R}$ and $\{L_i\}\subseteq \mathcal{R}$, we have that
  this set of assignments is identical for each $M_i$ and $M'_i$. Thus
  they must agree on the truth of $\varphi$, and therefore $\varphi$
  is also a discriminator for $\{M'_i\}$.\qed
\end{proof*}



\subsection{{\sc CLEVR} VDP Dataset}
\label{app:clevrdataset}
We describe the process of generating variants. Each base puzzle has a \emph{groundtruth} configuration that identifies the position and material attributes for each object in the scene. To obtain a variant for a given puzzle, we applied a transformation to the groundtruth configurations of all the images in the puzzle. These transformations would yield \emph{isomorphic} FO models.

Informally, an isomorphism in this our is a transformation that maps objects and relation symbols in a consistent and \emph{invertible} fashion. A simple example is that of uniformly replacing cubes with spheres and the attribute red with the attribute metallic in every image. Of course, one has to ensure that such a transformation is invertible, so if one replaced red items with the corresponding green items in an image with only red and green shapes, then the resulting image would only have green shapes and we would not be able to recover the original image using any transformation. One can show easily that such isomorphic transformations on puzzles with unique solutions will yield variants with unique solutions.

Second, we \emph{swap} the intended candidate image with one of the example images. We then check using the technique described in Appendix~\ref{app:implementation} that this swapped puzzle continues to have a unique (and in fact the same) discriminator, with the solution being the example image that was swapped in for the original candidate image. These are called \emph{allowable swaps}, and for a puzzle with 3 example images (as is the case with our {\sc Clevr} VDP Dataset) there are upto 3 allowable swaps since the intended candidate image can be swapped with any of the training images. Such allowable swaps are computed for each base puzzle and one is then applied at random.

Finally, we randomly jitter the positions of each object in every image in the variant configuration obtained by the isomorphic transformation to obtain the final configurations. These are then rendered into images using {\sc blender}.

\subsection{Details of Implementation}
\label{app:implementation}

\paragraph{Code for pretrained model} The code for the pretrained model provided by the authors of~\cite{nsvqa} can be found at: https://github.com/kexinyi/ns-vqa

\paragraph{Extracting FO Models} We use {\sc yolo}v4~\cite{yolov4} for the first phase on the Natural Scenes Dataset. For each image, the model provides a list of objects that are identified by class labels and bounding boxes. We then extract First-Order models from this representation by computing proxies for relationships such as \emph{left} or \emph{sitting on} by using the bounding boxes. We use the labels as unary predicates over the universe of objects, which then appear as guards in FO-SL. 

Similarly, for the {\sc Clevr} VDP Dataset we use a pretrained model from the work in~\cite{nsvqa}. The model returns a set of objects identified by attributes such as shape, color, or size, along with co-ordinate positions in 3D space (with respect to a fixed predefined origin and axis orientations). We then use treat the values for shape, color, etc as unary predicates (say $\mathit{red}(.)$ or $\mathit{large}(.)$) and compute spatial relationships such as $\mathit{left}(.,.)$ or $\mathit{behind}(.,.)$ between objects using their co-ordinates. Finally, we use the labels as guards in the FO-SL fragment corresponding to this logic.

\paragraph{SAT-based Synthesis} The SAT-based synthesis formulation for our symbolic synthesis engine is given in the main paper. Since discriminators are represented by boolean variables, for any given discriminator it is possible to negate the corresponding valuations of the boolean variables and add it as an \emph{additional constraint} to the SAT solver. This new SAT query would yield a discriminator that is different from the previous one. We used this technique to do an exhaustive search and ensure that the puzzles in the {\sc Clevr} VDP dataset have unique discriminators.

\subsection{Evaluation on Natural Scenes Dataset}
\label{app:naturaldatasetevaluation}
\begin{table*}
  \centering
  \begin{tabular}{lrrc}
    \toprule
    {\bf Target discriminator} & {\bf Sample Discriminators}
    \\
    \midrule
    All teddy bears on a sofa  & $\exists x.
    \,(\mathsf{sofa}(x)\wedge~\forall y.
    \,(\mathsf{teddy~bear}(y)\Rightarrow~\mathsf{sitting~on}(y,x)))$   \\
     & $\forall x.\,(\mathsf{teddy~bear}(x)\Rightarrow~\exists y.\,(\mathsf{sofa(y)}\wedge~\mathsf{sitting~on}(x,y)))$   \\

     \midrule
    There is an SUV  & $\exists x.\,\mathsf{truck}(x)$ \\

     \midrule
    Onward lane & $\forall x.\,(\mathsf{person}(x)\Rightarrow~\exists y.\,(\mathsf{car}(y)\wedge~\mathsf{left}(y,x)))$ \\
    (pedestrians to the right of traffic) & $\forall x.\,(\mathsf{person}(x)\Rightarrow~\forall y.\,(\mathsf{car}(y)\Rightarrow~\mathsf{right}(x,y)))$   \\

    \midrule
    Fruit in separate piles  & $\forall x.\,(\mathsf{orange}(x)\Rightarrow~\forall y.\,(\mathsf{apple}(y)\Rightarrow~\mathsf{left}(y,x)))$ \\

    \midrule
    Kickoff position & $\forall
                                 x.\,(\mathsf{person}(x)\Rightarrow~\forall
                                 y.\,(\mathsf{sports~ball}(y)\Rightarrow~\exists
                                 z.\,(\mathsf{person}(z)$
                                  \\
    (football in between two people)  &
                                              $\wedge~\mathsf{left}(y,x) \land \mathsf{right}(y,z))))$ \\

    \midrule
    Laid out place setting & $\forall x.\,(\mathsf{spoon}(x)\Rightarrow~\forall y.\,(\mathsf{fork}(y)\Rightarrow~\mathsf{left}(y,x)))$ \\
      & $\exists x.\,\mathsf{wine glass}(x)$ \\

    \midrule
    Person kicking a ball & $\exists x.\,(\mathsf{person}(x)\wedge~\exists y.\,\mathsf{sports~ball}(y))$ \\

    \midrule
    Dog herding sheep & $\exists x.\,(\mathsf{dog}(x)\land~\exists y.\,\mathsf{sheep}(y))$ \\

    \midrule
    Parking spot & $\exists x.\,\mathsf{parking~meter}(x)$ \\

    \midrule
    People carrying umbrellas & $\exists x.\,\mathsf{umbrella}(x)$ \\

    \midrule
    Bus with people in it & $\exists x.\,(\mathsf{bus}(x)\wedge~\exists y.\,(\mathsf{person}(y)\wedge~\mathsf{inside}(y,x)))$ \\
      & $\exists x.\,(\mathsf{person}(x)\wedge~\exists y.\,(\mathsf{bus}(y)\wedge~\mathsf{left}(y,x)))$ \\

    \midrule
    All dogs on sofas & $\forall x.\,(\mathsf{dog}(x)\Rightarrow~\exists y.\,(\mathsf{sofa}(y)\wedge~\mathsf{sitting~on}(x,y)))$ \\

    \midrule
    Desktop computer & $\exists x.\,(\mathsf{mouse}(x)\wedge~\exists y.\,(\mathsf{tvmonitor}(y)))$ \\
      & $\exists x.\,(\mathsf{tvmonitor}(x)\wedge~\exists y.\,\mathsf{keyboard}(y))$ \\
      & $\exists x.\,\mathsf{mouse}(x)$ \\

    \midrule
    People wearing ties & $\forall x.\,(\mathsf{tie}(x)\Rightarrow~\exists y.\,(\mathsf{person}(y)\wedge~\mathsf{wearing}(y,x)))$ \\
      & $\exists x.\,\mathsf{tie}(x)$ \\

    \midrule
    Person sleeping on a bench & $\exists x.\,\mathsf{person}(x)$ \\
     (versus animals sleeping) &  \\

    \midrule
    All cats on sofas & $\forall x.\,(\mathsf{sofa}(x)\Rightarrow~\forall y.\,(\mathsf{cat}(y)\wedge~\mathsf{sitting~on}(y,x)))$ \\
      &  $\exists x.\,(\mathsf{sofa}(x)\Rightarrow~\forall y.\,(\mathsf{cat}(y)\wedge~\mathsf{sitting~on}(y,x)))$ \\

    \midrule
    Kitchen & $\exists x.\,\mathsf{oven}(x)$ \\

    \midrule
    Two cats on the same sofa & $\exists x.\,(\mathsf{cat}(x)\wedge~\exists y.\,(\mathsf{sofa}(y)\wedge~\exists z.\,(\mathsf{sofa}(z)\wedge~\mathsf{left}(z,x))))$ \\

    \midrule
    Cat displayed on TV & $\exists x.\,(\mathsf{cat}(x)\wedge~\forall y.\,(\mathsf{tvmonitor}(y)\Rightarrow~\mathsf{displayed~on}(x,y)))$ \\
    \midrule
    TV is on & -- \\

    \bottomrule
  \end{tabular}
      \caption{Evaluation of framework on Natural Scenes Dataset ({\natnum} Visual Discrimination Puzzles total)}
      \label{largeyolotable}
\end{table*}

Table~\ref{largeyolotable} expands our reporting of the evaluation described in the main text. For each puzzle class, we give the English description of the intended discriminator and some examples of discriminators found in FO-SL by our tool.

\subsection{Details of the Neural Baseline solver for the {\sc Clevr} VDP Dataset}
\label{app:clevrsimilaritybaseline}
We detail the metric used for the baseline evaluation using image similarity.



We use a pretrained model of the similarity network based on the work in~\cite{wang2014learning} (code can be found at https://github.com/USCDataScience/Image-Similarity-Deep-Ranking). Given two images, this model computes a non-negative distance measure $d$.

Given a puzzle with example images $S_1, S_2, S_3$ and candidate images $C_1, C_2, C_3$ we compute the solution (choice of candidate) in the following manner:
\[
\mathsf{solution} = \mathsf{argmin}_{j\in [1,3]}\left(\prod\limits_{i=1}^{3}d(S_i, C_j)\right)
\]
i.e., we select the candidate whose product of distances from each of the training images is the least. 

The comparison of the efficacy of finding the target candidate by our tool and this baseline is depicted in Figure~\ref{fig:similaritybaseline}. 


\subsection{Details of the Prototypical Networks-based Baseline solver for the {\sc Clevr} VDP Dataset}
\label{app:clevrprototypebaseline}
We provide further detail about the Prototypical Networks~\cite{prototype} based baseline.

The idea behind a prototypical network is to learn an embedding of an image such that the `average' of several images belonging to a class behaves as the embedding of a prototypical member of that class. One can then use such prototypes computed from, say, a few positive and negative images to build a few-shot classifier. The classification is performed using a softmax over euclidean distances computed from the query point and the class prototypes.

We adapt this philosophy to develop a baseline for VDP. First initialise a ResNet-18~\cite{he2016deep} with pretrained weights from CIFAR 10. Then, we add a small MLP (linear + ReLU + linear) to the end of that network and `fine-tune' the weights of the entire network on 6 concept class schemas ($\sim$150 puzzles). The loss is computed as follows: given a puzzle with example images $s_1, s_2, s_3$ and candidate images $c_1, c_2, c_3$, we consider each candidate in turn to compute the score it achieves using the prototypical embedding and select the one with the maximum score. To compute the score, we consider that all the example images are in one class (say \emph{pos}), and each other candidate is in its own class (say \emph{neg1} and \emph{neg2}). We then compute the softmax score over distances with each of these `classes' for the candidate under consideration. The intuition, similar to the notion of discrimination, is that we compute the probability of the considered candidate belonging to the example set given that the other two candidates represent different ways of exhibiting non-membership in the example set. Formally:

\[
\mathsf{score}(c_i) = \frac{e^{-d(\mathit{prot}, c_i)}}{e^{-d(\mathit{prot}, c_i)} + \Sigma_{j = 1, j \neq i}^{3} e^{-d(c_j, c_i)}}
\]

where $\mathit{prot} = \frac{1}{3}\Sigma_{i=i}^{3}s_i$ is the prototype embedding computed from the example images, and $d$ is Euclidean distance.

We fine-tune with this score as the loss while validating on a held-out set of 4 schemas ($\sim$100 puzzles). We then select the model that had the best validation loss and evaluate it on the entire dataset. The results are presented in Figure~\ref{fig:similaritybaseline}. In developing this model, we found that this architecture worked better and somewhat consistently across all the concept classes. A similar architecture with a VAE initialised by pretraining on either {\clevr} or the VDP dataset itself overfit very quickly, and performed extremely well ($\sim$80\%) on a few concept classes while achieving 0\% on most others.

\subsection{Adapting Baselines for OddOne Dataset}
\label{app:oddonedatasetevaluation}
We adapt the baselines for the {\clevr} VDP dataset to the {\clevr} OddOne puzzle dataset. Recall that the puzzle calls for choosing the \emph{odd one} among a given set of images. Therefore, we choose the odd candidate by picking the one that maximises the distance to all other images. In the case of the image similarity baseline we use the similarity score given by the network as the inverse of distance, and in the case of the prototypical networks we use Euclidean distance computed over the learned embeddings.

\subsection{Bottom-Up Solver for Full First-Order Logic}
\label{app:bottomupsolver}
The bottom-up solver takes a set of positive- and negatively-labelled FO models and a grammar, and searches for a
classifier in the grammar that partitions the models (without using SAT solvers). 
It iterates from simpler to more complex formulas, heuristically growing the formula towards making it a discriminator.
In order to use this solver as a VDP solver, given a puzzle we create possible partitions by separating the images into the example images and one candidate image among the given candidates (positive), versus the rest of the candidates (negative). From our definition of a discriminator we know that any of the classifiers obtained thus would be discriminators, and the smallest and sensible discriminator would be the smallest sensible classifier among the possible classifiers across any partition. Therefore we extract FO models from the images in the puzzle, create partitions and use the bottom-up solver.

However, in practice if a particular partition is so malformed that the choice of candidate among the positive images is extremely incongruous with respect to the example images, then (especially for large grammars) it can take very long to exhaust the search space. Thus we run the solver for each possible choice of candidate image concurrently and return the first solution found, or wait until all runs fail to find a solution or times out. 

\subsection{Ablation Study on FO-SL}
\begin{table*}
  \centering
  \begin{tabular}{lll}
    \toprule
    {\bf Target Concept} & {\bf Sample Discriminators} & {\bf Natural
                                                         Language Reading}
    \\
     \\
    \midrule
    Two cats on a couch & $\exists x.~\forall y. \mathsf{cat}(x) \vee
                        \mathsf{right}(y,x)$ & \makecell[l]{``There is some
                                              cat or some thing \\ that all things are
                                              to the right of''} \\

    \midrule





    All cats are on sofas & $\forall x.~\exists
                            y. \neg \mathsf{chair}(x) \wedge \neg \mathsf{sofa}(y)$
                                                      & ``There is no chair and there is some non-sofa'' \\


    \midrule




    All dogs are on sofas & $\exists x.~\forall
                            y. \mathsf{sitting~on}(x,y) \vee
                            \neg\mathsf{sofa}(y)$  & ``There is
                                                     something that is sitting on all sofas'' \\

                         & $\exists x.~\exists
                                  y. \mathsf{sitting~on}(y,x) \wedge
                                  \neg\mathsf{dog}(y)$  & ``There is a
                                                          non-dog sitting on something'' \\

    \midrule
    Cat on TV & $\forall x.~\forall y. \mathsf{right}(x,y) \vee
                \neg\mathsf{displayed~on}(x,y)$ & \makecell[l]{``For every pair of
                                            things, either one is to
                                            the right \\ of the other or else
                                            it is not within the other''}  \\
                         & $\exists x.~\forall y. \mathsf{dog}(x) \vee
                          \mathsf{tvmonitor}(y)$ & ``Either there is
                                                    a dog or everthing
                                                    is a tv monitor'' \\

                         & $\exists x.~\forall y. \mathsf{displayed~on}(x,y)
                          \vee \neg\mathsf{tvmonitor}(y)$  & ``There
                                                              is
                                                              something
                                                              displayed on all tv monitors''\\

    \midrule
                      Desktop computer  & $\forall x.~\forall
                                          y. \neg\mathsf{sofa}(x)
                                          \wedge
                                          \neg\mathsf{placed~on}(x,y)$
                                                      & \makecell[l]{``There is no
    sofa and \\ there is nothing placed on anything else''} \\

    \midrule
    Dog herding & $\exists x.~\forall y. \mathsf{truck}(x) \vee
                  \neg\mathsf{person}(y)$ & ``There is a truck or
                                            there is no person''\\
                         & $\forall x.~\exists y. \mathsf{right}(x,y)
                          \vee \neg\mathsf{sheep}(x)$ & \makecell[l]{``Every
                                                         sheep has
                                                         something \\ that
                                                         it is to the
                                                         right of''}  \\

    \midrule

    Driving direction & $\forall x.~\exists y. \mathsf{left}(y,x) \vee
                        \neg\mathsf{car}(x)$ & ``Every car has
                                              something to its left''  \\
                         & $\forall x.~\exists y. \mathsf{truck}(y)
                          \vee \neg\mathsf{car}(x)$ & ``There is a
                                                      truck or there
                                                      is no car''  \\

    \midrule

    Kitchen & $\exists x.~\exists y. \mathsf{banana}(y) \vee
              \mathsf{oven}(x)$ & ``There is either a banana or an oven''   \\

                         & $\forall x.~\exists y. \mathsf{left}(x,y)
                          \vee \neg\mathsf{cup}(x)$ & ``Every cup is
                                                      to the left of something''  \\

    \midrule

    Parking meter in sight & $\forall x.~\exists y. \mathsf{left}(x,y)
                             \vee \mathsf{parkingmeter}(y)$ & \makecell[l]{``There
                                                              is a
                                                              parking
                                                              meter or
                                                              else \\ everything is to the left of something''} \\

    \midrule

    Person and ball & $\forall x.~\forall
                      y. \mathsf{person}(x) \vee
                      \neg\mathsf{had~by}(y,x)$ &
                                                  \makecell[l]{``Everything
                                                  is either a person \\ or else it has nothing''} \\

     \midrule
    Bus with people inside & $\exists x.~\exists y. \mathsf{bus}(x)
                             \wedge \mathsf{right}(x,y)$ & ``There is
                                                          a bus to
                                                          the right
                                                          of something'' \\
                         & $\exists x.~\exists y. \neg\mathsf{bus}(x)
                          \wedge \neg\mathsf{inside}(x,y)$ & \makecell[l]{``There
                                                              is a non-bus
                                                              and
                                                              something \\
                                                              that it is
                                                              not
                                                              inside''} \\

  \midrule

    People and ties & $\forall x.~\exists
                      y. \mathsf{tie}(y) \vee
                      \mathsf{wearing}(x,y)$ &
                                              \makecell[l]{``There
                                              is
                                              a
                                              tie
                                              or else \\ everything is wearing something''} \\

    \midrule

    People wearing ties & $\exists x.~\forall
                          y. \mathsf{right}(x,y) \vee
                          \mathsf{tie}(x)$ & \makecell[l]{``There is a tie or else
                                             there is \\ something to the
    right of everything''} \\


    \midrule
    Someone sleeping on a bench & $\exists x.~\forall
                                  y. \neg\mathsf{left}(y,x) \wedge
                                  \neg\mathsf{on}(y,x)$ & \makecell[l]{``There
                                                              is
                                                              something
                                                              with
                                                              nothing
                                                              to the
                                                              left of
                                                              it \\ and nothing on it''} \\

                         & $\exists x.~\exists y. \mathsf{right}(y,x)
                          \wedge \neg\mathsf{bench}(x)$ & ``There is
                                                          something
                                                          to the
                                                          right of
                                                          a non-bench'' \\

    \midrule

    Umbrella weather & $\exists x.~\forall y. \mathsf{umbrella}(x)
                      \vee \mathsf{had~by}(x,y)$ & \makecell[l]{``There is an
                                                    umbrella or else \\
                                                    there is something
                                                    had by everything''} \\

    \bottomrule
  \end{tabular}
      \caption{Discriminators found \emph{without} restriction to FO-SL which end up being unnatural}
      \label{nonguardedtablelarge}
\end{table*}

We provide further details of the ablation study on FO-SL. Recall that the study was performed by replacing the symbolic synthesis engine that searches for discriminators over FO-SL with another solver that searches for discriminators in full First-Order Logic by building formulas in a bottom-up manner (details of this bottom-up solver are given below). The discriminators returned by this solver need not have guarded quantifiers or conjunctions.

A complete description of the discriminators found by this solver on the Natural Scenes VDP Dataset is given in Table~\ref{nonguardedtablelarge}. As is evident, the symbolic synthesis picks on many discriminators that are unnatural--- unguarded quantification leads to talking about ``things'' rather than more natural kinds of objects and disjunction leads to picking on weird coincidences that happen to tie images together (like the discriminator found on the class ``All cats are on sofas'' being ``There is no chair and there is some non-sofa'', or ``Desktop computer'' being ``There is no sofa and there is nothing placed on anything else''). A key observation here is that it is easy to tie almost any set of few (say 4) images with a concept that involves disjunction, and hence disjunctive concepts do not pick up on true similarity between images.

\end{document}